%% file: amass.tex
\documentclass[10pt,twocolumn,letterpaper]{article}

\usepackage{iccv}
\usepackage{times}
\usepackage{epsfig}
\usepackage{graphicx}
\usepackage{amsmath}
\usepackage{amssymb}
\usepackage{xcolor}
\usepackage{booktabs} %

\usepackage{afterpage}
\usepackage[]{subfig}

\input{notation}
\usepackage{balance}

\usepackage[pagebackref=true,breaklinks=true,letterpaper=true,colorlinks,bookmarks=false]{hyperref}

\iccvfinalcopy %

\newif\ifuseDLExperiment

\ificcvfinal\fi
\begin{document}

\title{AMASS: Archive of Motion Capture as Surface Shapes}

\author{Naureen Mahmood\\
Meshcapade GmbH\\
Karlstrasse 3\\
72072 T\"{u}bingen, Germany\\
{\tt\small naureen@meshcapade.com}
\and
Nima Ghorbani\\
MPI for Intelligent Systems\\
Max-Planck-Ring 4\\
72076 T\"{u}bingen, Germany\\
{\tt\small nima.ghorbani@tue.mpg.de}
\and
Nikolaus F. Troje\\
York University\\
4700 Keele St, Toronto\\ 
ON M3J 1P3, Canada\\
{\tt\small troje@yorku.ca}
\and
Gerard Pons-Moll\\
MPI for Informatics \\
 Saarland Informatics Campus\\
Campus E1 4, Stuhlsatzenhausweg\\
66123 Saarbr\"{u}cken, Germany\\
{\tt\small gpons@mpi-inf.mpg.de}
\and
Michael J. Black\\
MPI for Intelligent Systems\\
Max-Planck-Ring 4\\
72076 T\"{u}bingen, Germany\\
{\tt\small black@tue.mpg.de}
}

\maketitle

\begin{abstract}
\input{sections/00_abstract}

\end{abstract}

\input{sections/01_intro}

\input{sections/02_previous}

\input{sections/03_technichal}

\input{sections/08_evaluation}

\input{sections/09_amass}
\input{sections/10_future}

{\small
	\balance
	\bibliographystyle{ieee}
	\bibliography{egbib}
}

\newpage
\part*{APPENDIX}
\setcounter{section}{0}
\input{sections/11_acks}

{\bf Conflict of Interest Disclosure:} 
N. Mahmood is a founder and shareholder of Meshcapade GmbH, which is commercializing body shape technology. N. Mahmood is with Meshcapade GmbH but this work was performed primarily at the MPI for Intelligent Systems. Michael J. Black has received research gift funds from Intel, Nvidia, Adobe, Facebook, and Amazon. While MJB is a part-time employee of Amazon, his research was performed solely at, and funded solely by, MPI. 
MJB is also an investor in Meshcapde GmbH.

\input{sections/supplementary/optimization}
\input{sections/supplementary/ssm_hardware}
\input{sections/supplementary/model_size}

\input{sections/supplementary/video_samples}

\end{document}

%% file: notation.tex
\newcommand{\R}{\mathbb{R}}      %

\newcommand{\amassNDatasets}{15}
\newcommand{\amassHoursApprox}{40}
\newcommand{\amassHoursExact}{42}

\newcommand{\amassNSubjApprox}{300}
\newcommand{\amassNSubjExact}{346}
\newcommand{\amassNMotionsApprox}{11000}
\newcommand{\amassNMotionsExact}{11451}
\newcommand{\amassWebAddress}{http://amass.is.tue.mpg.de}

\newcommand{\numBetas}{16}
\newcommand{\numDMPLs}{8}
\newcommand{\numPose}{90}

\newcommand{\numMANOBH}{24}
\newcommand{\numMANOFull}{90}
\newcommand{\vect}[1]{\mathbf{#1}}
\newcommand{\mat}[1]{\vect{#1}}

\newcommand{\shape}{{\boldsymbol{\beta}}}
\newcommand{\pose}{{\boldsymbol{\theta}}}

\newcommand{\dyna}{{\boldsymbol{\phi}}}

\newcommand{\Poses}{\Theta}

\newcommand{\bweights}{\mat{W}}
\newcommand{\joints}{\mat{J}}
\newcommand{\posebs}{B_p(\pose)}          %
\newcommand{\shapebs}{B_s(\shape)}          %

\newcommand{\vt}{\mat{T}_\mathrm{\mu}}                        %

\newcommand{\Obsmrks}{\mathcal{M}}
\newcommand{\obsmrk}{\mathbf{m}}
\newcommand{\latentmrk}{\tilde{\mathbf{m}}}
\newcommand{\Latentmrks}{\tilde{\mathcal{M}}}

\newcommand{\mrkmap}{m}

\newcommand{\mrkDistance}{9.5mm}
\newcommand{\StoM}{R}

\newcommand{\StoMwt}{\lambda_\StoM}
\newcommand{\shapewt}{\lambda_{\shape}}
\newcommand{\initwt}{\lambda_I}
\newcommand{\datawt}{\lambda_D}
\newcommand{\posewt}{\lambda_{\pose}}
\newcommand{\dynawt}{\lambda_{\dyna}}

\newcommand{\poseConstwt}{\lambda_u}
\newcommand{\dynaConstwt}{\lambda_v}

\newcommand{\poseConst}{u}
\newcommand{\dynaConst}{v}
\newcommand{\posecpl}{q}

\newcommand{\dataset}{SSM} %

%% file: sections/00_abstract.tex
Large datasets are the cornerstone of recent advances in computer vision using deep learning.
In contrast, existing human motion capture (mocap) datasets are small and the motions limited, hampering progress on learning models of human motion.
While there are many different datasets available, they each use a different parameterization of the body, making it difficult to integrate them into a single meta dataset. 
To address this, we introduce AMASS, a large and varied database of human motion that unifies $\amassNDatasets$ different optical marker-based mocap datasets by representing them within a common framework and parameterization. 
We achieve this using a new method, MoSh++, that converts mocap data into realistic 3D human meshes represented by a rigged body model; here we use SMPL~\cite{SMPL:2015}, which is widely used and provides a standard skeletal representation as well as a fully rigged surface mesh.
The method works for arbitrary marker sets, while recovering soft-tissue dynamics and realistic hand motion.
We evaluate MoSh++ and tune its hyperparameters using a new dataset of 4D body scans that are jointly recorded with marker-based mocap.
The consistent representation of AMASS makes it readily useful for animation, visualization, and generating training data for deep learning.
Our dataset is significantly richer than previous human motion collections, having more than $\amassHoursApprox$ hours of motion data, spanning over $\amassNSubjApprox$ subjects, more than $\amassNMotionsApprox$ motions, and will be publicly available to the research community.

%% file: sections/01_intro.tex
\section{Introduction}\label{sec:introduction}
\begin{figure}[t]
	\begin{center}
	\includegraphics[width=3.1in]{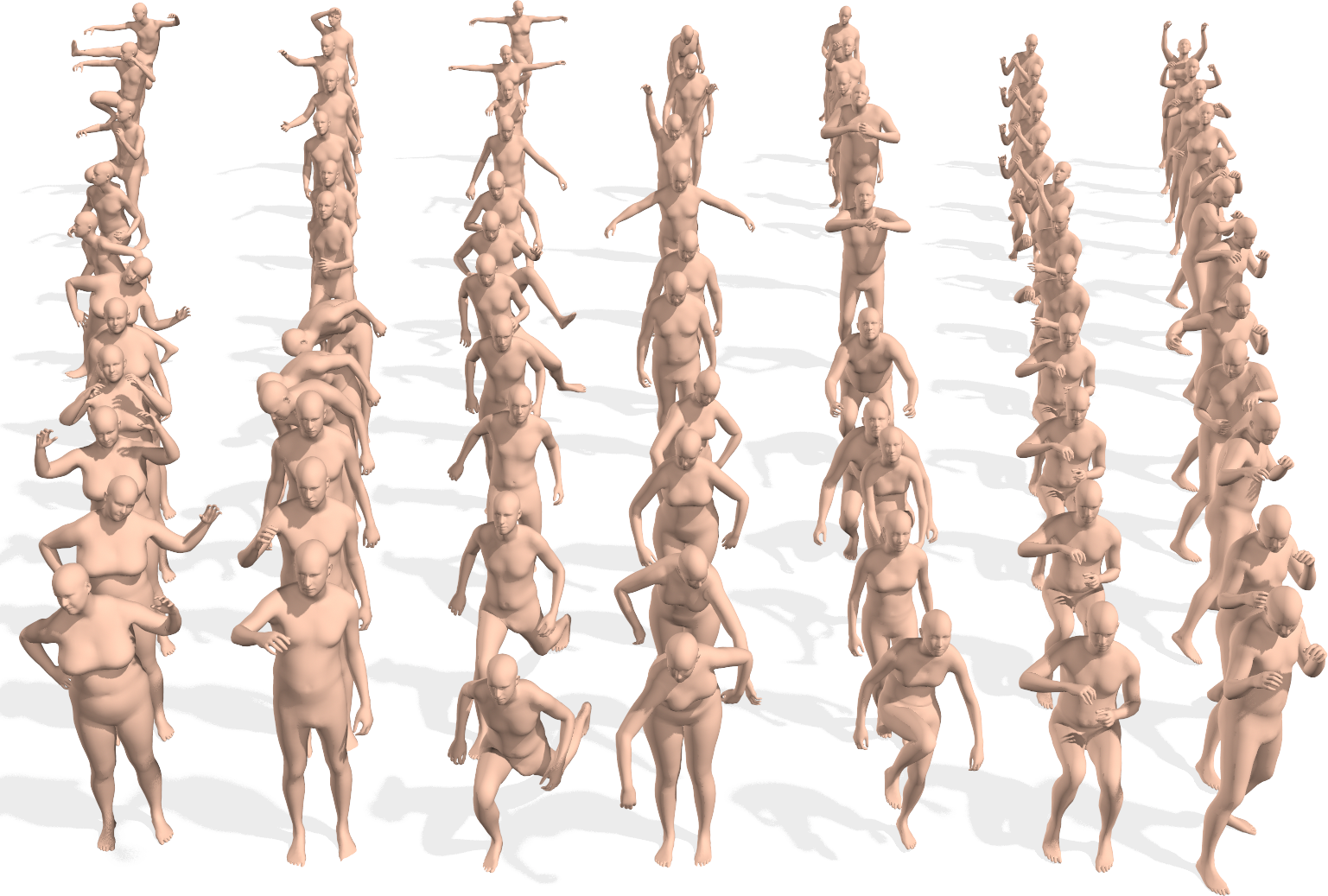}
	\end{center}
	\caption{We unify a large corpus of archival marker-based optical human mocap datasets by 
			representing them within a common framework and parameterization. A sampling of shapes and poses from a few
			datasets in AMASS is shown, from left to right: CMU \cite{CMU}, MPI-HDM05 \cite{MPI_HDM05,Mueller_09},
			MPI-Pose Limits \cite{MPI_limits}, KIT \cite{Mandery2015_KITMotionDB}, BioMotion Lab \cite{BioMotion}, TCD \cite{TCD_hands} and ACCAD \cite{ACCAD} datasets. The input
			is sparse markers and the output is SMPL body models.}
	\label{fig:amass_small}
\end{figure}
\begin{figure}[!t]
	\begin{center}
		\includegraphics[width=3.1in]{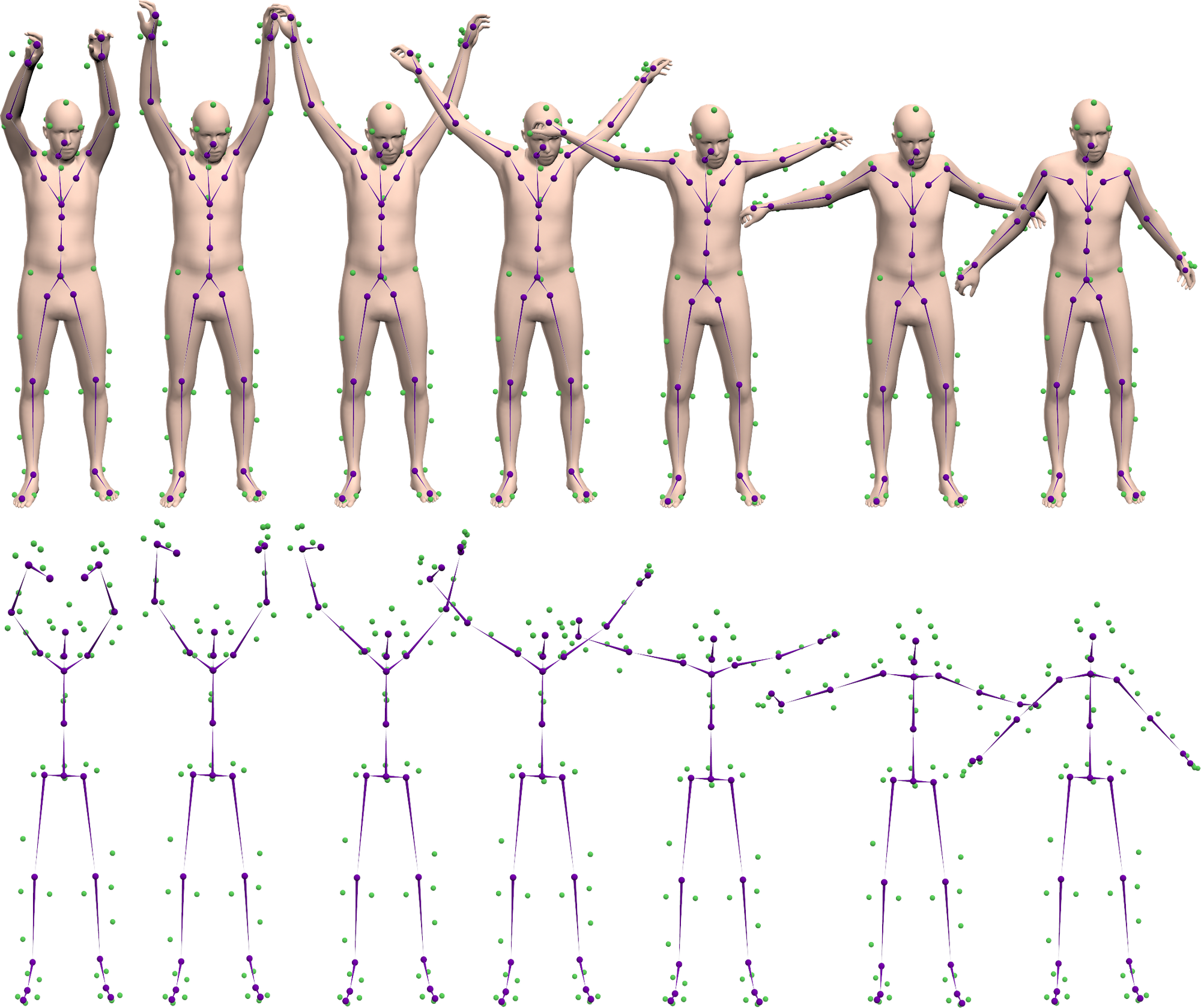}
	\end{center}
	\caption{MoSh++ captures body shape, pose, and soft-tissue dynamics by fitting the surface of the SMPL/DMPL body model to observed mocap markers,
		 	while also providing a rigged skeleton that can be used in standard animation programs (top row).
		  	Conventional mocap methods estimate skeletal joints from the markers, 
		  	filtering out surface motion as noise and losing body shape information (bottom row). Original mocap markers are shown in green.}
  \label{fig:mocap_vs_mosh}
\end{figure}
This paper addresses two interrelated goals.
First, we develop a method to accurately recover the shape and pose of a person in motion from standard motion capture (mocap) marker data.
This enables the second goal, which is to create the largest publicly available database of human motions that can enable machine learning for applications in animation and computer vision.
While there have been attempts in both of these directions, existing mocap databases are insufficient in terms of size and complexity to exploit the full power of existing deep learning tools.
There are many different mocap datasets available, but pulling them together into a coherent formulation is challenging due to the use of widely varying markersets and laboratory-specific procedures \cite{gorton2009assessment}.
We achieve this by extending MoSh \cite{loper2014mosh} in several important ways,  enabling us to collect a large and varied dataset of human motions in a consistent format (Fig.~\ref{fig:amass_small}).

MoSh employs a generative model of the body, learned from a large number of 3D body scans, to compute the full 3D body shape and pose from a sparse set of motion capture markers.
The results are realistic, but the method has several important limitations, which make it inappropriate for our task.
First, MoSh relies on a formulation of the {\em SCAPE body model} \cite{Anguelov05}, which is not compatible with existing body representations and graphics software, making it a poor choice for distributing a dataset. 
We replace SCAPE with the SMPL body model \cite{SMPL:2015}, which uses a kinematic tree, has joints, and is based on blend skinning. 
SMPL comes with a UV map, which allows researchers to generate their own textures for rendering images and video sequences.
SMPL is readily available, widely used, and compatible with most game engines and graphics packages.
Second, while MoSh captures some soft tissue motions, these are approximate and represented by changing the {\em identity} of a subject over time; that is, they are not true soft-tissue deformations.
Here we take the dynamic shape space from DMPL~\cite{SMPL:2015}, which models these soft-tissue deformations from the SMPL model using a shape space learned from 4D body scans of various subjects. 
We show that we can recover the soft tissue motions realistically from a sparse set of markers. The resulting body shapes and motions look natural and we show that they are metrically accurate.
Third, MoSh did not solve for the pose and motion of the hands.
Here we add the recent MANO hand model~\cite{MANO:SIGGRAPHASIA:2017}, which is compatible with SMPL, and solve for body and hand pose when hand markers are present.
This provides richer and more natural animations.    
Fourth, to fine tune and evaluate our proposed method, we collect a novel dataset, {\em Synchronized Scans and Markers (SSM)}, that consists of dense 3D meshes in motion, captured with a 4D scanner, together with traditional marker-based mocap. We separate the sequences into training and testing sets, and train the hyperparameters of MoSh++ to minimize the distance between the ground truth 3D scans and the estimated 3D body meshes. We then evaluate the performance of MoSh++ on the test set, demonstrating the accuracy of the method and allowing a quantitative comparison to MoSh.

MoSh++ enables our key goal of creating a large database of human motions.
While there are many motion capture datasets available online for research purposes \cite{MPI_limits,TotalCapture,loper2014mosh,CMU,HEva,CMUKitchen,BioMotion,ACCAD,TCD_hands,MPI_HDM05}, they are often narrowly focused on a particular class of motions.
Even the largest ones are too limited in size and variety to support serious deep learning models.
Additionally, datasets vary in the format of the data and the kinematic structure of the body, making it hard for researchers to combine them.
There have been several efforts to create data supersets \cite{MocapClub,holden2016deep,Mandery2015_KITMotionDB}, but the process of unifying the datasets typically means standardizing to fixed body proportions, which fundamentally alters the data.
We argue that a good dataset should capture the articulated structure of the body in a way that is consistent with standard body models so that it can easily be adapted to new problems.
Additionally, richness of the original source marker data should be retained as much as possible.
It should also be possible to produce high-quality animations that are realistic enough to train computer vision algorithms; that is, the dataset should include full 3D human meshes. 

SMPL provides the unifying representation that is independent of the marker set, yet maintains the richness of the original marker data, including the 3D body shape.
We know of no other attempt that provides access to full body shape and soft-tissue from mocap data, while also providing accurate body and hand pose.
Here we combine $\amassNDatasets$ existing motion capture datasets into one large dataset: the {\em Archive of Mocap as Surface Shapes (AMASS)}.
AMASS has $\amassHoursExact$ hours of mocap, $\amassNSubjExact$ subjects, and $\amassNMotionsExact$ motions.
The source datasets all contain varying markersets ranging in size from 37 to 91 markers; AMASS unifies these into a single format. 
Each frame in AMASS includes the SMPL 3D shape parameters ($\numBetas$ dimensions), the DMPL soft tissue coefficients ($\numDMPLs$ dimensions), and the full SMPL pose parameters ($\numPose$ dimensions), including hand articulations. 
Users who only care about pose can ignore body shape and soft tissue deformations if they wish. 
Similarly, the SMPL shape space makes it trivial to normalize all bodies to the same shape if users want joint locations normalized to a single shape.
Fig. \ref{fig:amass_small} shows a selection of poses and body shapes in the dataset while
Fig. \ref{fig:mocap_vs_mosh} illustrates the difference between MoSh++ and traditional mocap.
Traditional datasets contain skeletons and/or makers, while the AMASS dataset also provides fully rigged 3D meshes.
With MoSh++  it is easy to add more data and we will continue to expand the dataset.
We make AMASS available to the research community at \url{\amassWebAddress}, and will support the community in adding new captures as long as they can be similarly shared.

In summary, we provide the largest unified mocap dataset (AMASS) to the community, enabling new applications that require large amounts of training data.

%% file: sections/02_previous.tex
\section{Related Work}
There is a large literature on estimating skeletal parameters from mocap markers as well as several commercial solutions that solve this problem.
As shown by Gorton et al.~\cite{gorton2009assessment}, different solutions use different skeletal models and pre-specified markersets, which makes it hard to unify the existing corpora of marker-based human recordings. Furthermore, all the methods that fit skeletons to data effectively lose rich surface information in the process. 
We review the most related work: fitting surface models to markers, %
capturing hands and soft-tissue motion from markers,  %
and previous motion capture datasets.

{\bf Surface Models from Markers.}
To reconstruct bodies from markers, most methods first build a statistical model of body shape \cite{allen2003space} or body shape and pose \cite{Allen:2006:LCM,Anguelov05,SMPL:2015}.
Allen et al.~\cite{allen2003space} reconstruct body shape using $74$ landmarks. 
They do this only for a fixed body pose and the correspondences between the model and the markers are known. 
The approach cannot deal with arbitrary poses because the model cannot be posed.
Anguelov et al.~\cite{Anguelov05} go further by learning a model (SCAPE) of shape and non-rigid pose deformations. Their method requires a dense 3D scan of each subject, which restricts its application to archival mocap datasets of markers.

Loper et al.~\cite{loper2014mosh} address some of these limitations with MoSh, and remove the requirement for individual 3D dense scans. 
MoSh uses a BlendSCAPE body model formulation \cite{hirshberg2012coregistration}, which is not compatible with standard graphics packages making it sub-optimal for distribution. Furthermore, MoSh does not capture real soft-tissue dynamics, and does not capture hands.

{{\bf Hands. }}
There is a large body of work on fitting hand models to RGB-D data~\cite{taylor2016efficient,tkach2017online} but here we focus on methods that capture hand motion from sparse markers. 
Maycock et al.~\cite{maycock2015fully} combine an optimal assignment method with model fitting but can capture only hands in isolation from the body and require a calibration pose. 
Schroder et al.~\cite{schroder2015reduced} propose an optimization method to find a reduced sparse marker set and, like us, they use a kinematic subspace of hand poses.
Alexanderson et al.~\cite{alexanderson2016robust} capture hand motion using sparse markers (3-10). They generate multiple hypotheses per frame and then connect them using the Viterbi algorithm \cite{forney1973viterbi}. 
They can track hands that exit and re-enter the scene and the method runs in real time. However, a new model needs to be trained for every marker set used.
Han et al.~\cite{Han:2018:OOM} address the problem of automatically labeling hand markers using a deep network.
The above methods, either do not estimate hands and bodies together or do not provide a 3D hand shape.

{{\bf Soft-tissue motion.}}
Most of the work in the mocap community focuses on \emph{minimizing} the effect of skin deformations on the marker motions~\cite{ANDRIACCHI20001217,leardini2005human}. In some biomechanical studies, the markers have even been fixed to the bones via percutaneous pins~\cite{lafortune1992three}. Our work is very different in spirit. We argue that, for animation, such soft-tissue and skin deformation makes captured subjects look alive. In \cite{loper2014mosh} they capture soft-tissue by fitting the parameters of a space of static body shapes to a sparse set of markers. 
This corresponds to modeling soft-tissue deformation by changing the {\em identity} of a person.
Instead, inspired by Dyna~\cite{Dyna:SIGGRAPH:2015}, we use the dynamic shape space of DMPL~\cite{SMPL:2015}, which is learned from thousands of dynamic scans. This results in more realistic soft-tissue motions with minimal increase in model complexity.%

{\bf Motion Capture Datasets.}
There are many motion capture datasets \cite{TotalCapture,CMU,Mueller_09,MPI_HDM05,HEva,MPI_limits,BioMotion,Gymnasts} as well as several attempts to aggregate such datasets into larger collections \cite{MocapClub,holden2016deep,Mandery2015_KITMotionDB}.
Previous attempts to merge datasets \cite{holden2016deep,Mandery2015_KITMotionDB} adopt a common body representation in which the size variation among subjects is normalized.  
This enables methods that focus on modeling pose and motion in terms of joint locations.
On the other hand, such an approach throws away information about how body shape and motion are correlated and can introduce artifacts in retargeting all data to a common skeleton.
For example, Holden et al.~\cite{holden2016deep} retarget several datasets to a common skeleton to enable deep learning using {\em joint positions}.
This retargeting involves an inverse kinematics optimization that fundamentally changes the original data.

Our philosophy is different. %
We work directly with the markers and not the skeleton, recovering the full 3D surface of the body.
There is no loss of generality with this approach as it is possible to derive any desired skeleton representation or generate any desired markerset from the 3D body model.
Moreover, having a body model makes it possible to texture and render virtual bodies in different scenes.
This is useful for many tasks, including generating synthetic training for computer vision tasks  \cite{varol17}.

%% file: sections/03_technichal.tex
\section{Technical Approach}
To create the AMASS dataset, we generalize MoSh in several important ways:  1) we replace BlendSCAPE by SMPL to democratize its use (Sec.~\ref{sec:body_model}); %
2) we capture hands and soft-tissue motions (Sec.~\ref{sec:model_fitting});  
3) we finetune the weights of the objective function using cross-validation on a novel dataset, SSM (Sec.~\ref{sec:evaluation}). 
\input{sections/04_bodymodel}

\input{sections/model_fitting}

\input{sections/07_optimization}

%% file: sections/04_bodymodel.tex
\subsection{The Body Model}
\label{sec:body_model}

\ifuseDLExperiment
\else
	\begin{figure*}[!t]
		\centering{
			\includegraphics[width=\linewidth]{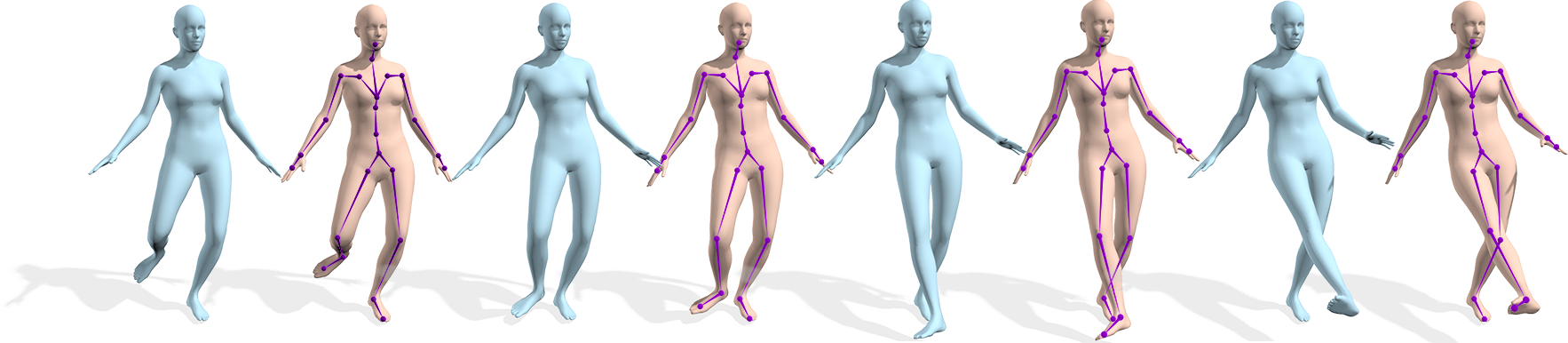} 
		}
		\vspace{-0.1in}
		\caption{ 
			MoSh and BlendSCAPE (blue) vs.~MoSh++ and SMPL (orange). These are visually similar, but MoSh++ is more accurate and SMPL provides a standard rigged mesh with a skeleton.
		}
		\label{fig:mosh_vs_moshpp}
	\end{figure*}
\fi

AMASS is distributed in the form of SMPL body model parameters.
SMPL uses a learned rigged template $\mat{T}$ with $N=6890$ vertices.
	The vertex positions of SMPL are adapted according to identity-dependent shape parameters, $\shape$, the pose parameters, $\pose$, and translation of the root in the world coordinate system, $\gamma$.
	The skeletal structure of the human body is modeled with a kinematic chain consisting of rigid bone segments linked by joints. 
	Each body joint has 3 rotational Degrees of Freedom (DoF), parametrized with exponential coordinates. 
    We use a variant of SMPL, called SMPL-H \cite{MANO:SIGGRAPHASIA:2017}, which adds hand articulation to the model using a total of $n=52$ joints, where $24$ joints are for the body and the remaining $28$ joints belong to the hands. 
    For simplicity of notation, we include the 3D translation vector $\gamma$ in the pose vector. 
The pose $\pose$ is determined by a pose vector of $3\times{52} + 3 = 159$ parameters.
    The remaining attributes of the SMPL-H model are the same as SMPL.
	
We combine SMPL-H with DMPL to obtain a model that captures both hand pose and soft-tissue deformations. %
For brevity we refer to the combined SMPL-H + DMPL model as SMPL throughout this paper, although this goes beyond any published model.

SMPL modifies the template in an additive way. It applies additive shape, pose, and dynamic blendshapes to a template in a canonical pose and predicts joint locations from the deformed surfaces.
	The model is 
    	\begin{eqnarray}
        	S(\shape, \pose, \dyna) & = & G
        	(T(\shape,\pose,\dyna),J(\shape),\pose,\bweights ) \\
           \label{eq:SMPL}
        	T(\shape, \pose, \dyna)
        	& = & \vt+\shapebs+\posebs + B_d(\dyna)
    	   \label{eq:SMPL_template}
    	\end{eqnarray}
    where $G(\mathbf{T}, \joints, \pose,\bweights) : \R^{3N}\times{\R}^{|\theta|}\times{\R}^{3K}\times{\R^{4\times{3N}}}\mapsto \R^{3N}$ is a
	linear blend skinning function that takes 
	vertices in the model in the rest pose $\mathbf{T}$,
 $K$ joint locations stacked in $\joints$, a pose $\pose$, and the
	blend weights $\bweights$, and returns the posed vertices. 
	The blendshape functions $\shapebs$, $\posebs$, and $B_d(\dyna)$ output vectors of 
	vertex offsets relative to the mean template, $\vt$ (refer to \cite{SMPL:2015}, \cite{Dyna:SIGGRAPH:2015} for the detailed explanation of the functions).
	We call these  shape, pose, and dynamic blend shapes respectively. 
 Note that the pose blendshape are a function of the pose $\pose$, while $\shape$ and $\dyna$ correspond to linear coefficients that determine the shape and soft-tissue deformation.
    
	SMPL captures the dimensionality of body space more compactly than BlendSCAPE.
With only  $\numBetas$ shape, and $\numDMPLs$ dynamics components, MoSh++ achieves better accuracy than MoSh using $100$ shape components. 
The number of shape and dynamics coefficients is chosen using the SSM dataset such that MoSh++ does not overfit to mocap markers (see Supplemental Material).

%% file: sections/model_fitting.tex
\subsection{Model Fitting} \label{sec:model_fitting}
Similar to MoSh~\cite{loper2014mosh}, MoSh++ uses two stages to fit a body model to a sparse markerset. In the following we explain these two stages, while briefly mentioning the overlapping parts for completeness, and highlighting our contributions. For comparability we use a similar notation to the original MoSh paper.

{\bf Stage I}: Following MoSh, we use a marker parametrization $\mrkmap(\latentmrk_i,\shape,\pose_t)$ that maps a latent, pose invariant representation of the markers, $\latentmrk_i$, to estimate their position in a posed frame, $\pose_t$. In the first stage, for $F=12$ randomly chosen frames from the subject-specific mocap sequences, given an initial guess for marker-body correspondences, we optimize poses $\Poses=\pose_{1\hdots F}$, a single shape $\shape$, and latent marker positions $\Latentmrks = \{\latentmrk_i\}$ to fit the observed marker locations $\Obsmrks=\{\obsmrk_{i,t}\in \Obsmrks_t\}_{1\hdots F}$, where $i$ indexes the markers in a frame; at this stage we exclude soft-tissue deformations. More specifically, similar to MoSh, we optimize the following objective function:
\begin{equation}
\begin{split}
E(\Latentmrks, \shape, \Poses_B, \Poses_H) 
&= \datawt E_D(\Latentmrks, \shape, \Poses_B, \Poses_H) 
+ \shapewt E_{\shape}(\shape)  \\
&+ {\posewt}_B E_{\pose_B}(\pose_B) 
+ {\posewt}_H E_{\pose_H}(\pose_H)     \\
&+ \StoMwt E_{\StoM}(\Latentmrks, \shape)  
+ \initwt E_I(\Latentmrks, \shape)   .
\label{eq:mosh_shape_obj}
\end{split}
\end{equation}
The data term $E_D$ measures distance between \emph{simulated markers} $\mrkmap(\latentmrk_i,\shape,\pose_t)$ and the observed ones $\obsmrk_{i,t}$; $E_{\shape}$ is a Mahalanobis distance shape prior on the SMPL shape components; $E_{\pose_B}$ regularizes the body pose parameters; $E_{\StoM}$ encourages the latent markers to remain a prescribed distance $d$ from the body surface (here we use an average value of $d=\mrkDistance$); and $E_I$ penalizes deviations of latent markers from their initialized locations defined by the markerset (see \cite{loper2014mosh} for further details). 

In addition to the original terms of MoSh in Eq. \ref{eq:mosh_shape_obj}, we add $E_{\pose_H}$, which regularizes the hand pose parameters.
We project the full hand pose (i.e. $\numMANOFull$ hand parameters) into the $\numMANOBH$-D MANO pose space for both hands and compute the Mahalanobis distance in this space
\begin{equation}
E_{\pose_H}(\pose_H) =
{\hat{\pose}_H}^T \Sigma_{\pose_H}^{-1} {\hat{\pose}_H},
\label{eq:mano_prior}
\end{equation}
where $\hat{\pose}$ represents the projection of the pose and $\Sigma_{\pose_H}$ is the diagonal covariance matrix of the low-D PCA coordinates (see ~\cite{MANO:SIGGRAPHASIA:2017}). 

In contrast to MoSh, the $\lambda$ hyper-parameters are determined by line search on the training set of SSM (Sec. \ref{subsec:gridsearch}).
The data term, $E_D$, in  Eq.~\ref{eq:mosh_shape_obj} uses a sum of squared distances, which is affected by the number of markers in the observed mocap sequence. This is noteworthy since a standard $46$-markerset was used to determine the $\lambda$ weights during the hyper-parameter search. 
To deal with varying numbers of markers, we automatically vary the weight of this term by scaling it  by a factor $b = 46/n$, where $n$ is the number of observed mocap markers.

To help avoid local optima while minimizing Eq.~\ref{eq:mosh_shape_obj}, we use Threshold Acceptance method \cite{dueck1990threshold} as a fast annealing strategy. Over $4$ annealing stages of graduated optimization, we increase $\datawt$ by multiplying it by a constant factor $s=2$ while dividing the regularizer weights by the same factor. The weights at the final iteration are as follows:
\begin{multline}
\datawt=600 \times b, 
\shapewt=1.25, 
{\posewt}_B=0.375,\\ 
{\posewt}_H=0.125,
\initwt=37.5, 
\StoMwt=1e4 
\end{multline}
The surface distance regularization weight, $\StoMwt$, remains constant throughout the optimization.  The $\numMANOBH$ hand pose components are added into the optimization only during the final two iterations.

{\bf Stage II}: %
In this stage, the latent marker locations and body shape parameters $\shape$ of the model are assumed constant over time and the objective at this stage optimizes pose for each frame of mocap in the sequence. %

Like MoSh, we add a temporal smoothness term for pose changes, $E_{\poseConst}$, to help reduce the effect of jitter noise in the mocap marker data.  
Yet in contrast to MoSh, we optimize for the soft-tissue deformation coefficients, $\dyna$. We add a prior and a temporal smoothness terms, $E_{\dyna}(\dyna)$ and  $E_{\dynaConst}(\dyna)$ respectively, to regularize the soft-tissue deformations.
Then the final objective function for this stage becomes
        \begin{equation}
         \begin{split}
            E(\pose_B, \pose_H, \dyna) 
            &= \datawt E_D(\pose_B, \pose_H, \dyna) \\
            &+ {\posewt}_B E_{\pose_B}(\pose_B) 
             + {\posewt}_H E_{\pose_H}(\pose_H) \\
            &+ \poseConstwt E_{\poseConst}(\pose_B, \pose_H) \\ 
            &+ \dynawt E_{\dyna}(\dyna) + \dynaConstwt E_{\dynaConst}(\dyna)  .
            \label{eq:mosh_pose_obj}
         \end{split}
        \end{equation}
The data, body, and hands pose prior terms, $E_D$, $E_{\pose_B}$, and $E_{\pose_H}$, are the same as described in the first stage. 
To regularize the soft-tissue coefficients, we add a Mahalonobis distance prior on the $\numDMPLs$ DMPL coefficients.
        \begin{equation}
            E_{\dyna}(\dyna) =
            \dyna_t^T \Sigma_{\dyna}^{-1} \dyna_t,
            \label{eq:dyna_prior}
        \end{equation}
where the covariance $\Sigma_{\dyna}$ is the diagonal covariance matrix from the DYNA dataset~\cite{Dyna:SIGGRAPH:2015}. 

When hand markers are present, MoSh++ optimizes the hand pose parameters in the same way as all the other pose parameters except that we use $\numMANOBH$ dimensions of MANO's \cite{MANO:SIGGRAPHASIA:2017} low-dimensional representation of the pose for both hands. In cases where there are no markers present on the hands of the recorded subjects, the hand poses are set to the average pose of the MANO model.

The initialization and fitting for the first frame of a sequence, undergoes a couple of extra steps compared to the rest of the motion. 
For the first frame, we initialize the model by performing a rigid transformation between the estimated and observed markers to repose the model from its rest pose $\mathbf{T}$ to roughly fit the observed pose. Then we use a graduated optimization for Eq.~\ref{eq:mosh_pose_obj} with only the data and body pose prior terms, while ${\posewt}_B$ is varied from $[10, 5, 1]$ times the final weight.
Later, for each of the subsequent frames, we initialize with the solution of the previous frame to estimate the pose and soft-tissue parameters.

The per-frame estimates of dynamics and pose after the first frame are carried out in two steps. During the first step, we remove the dynamics and dynamics smoothness terms, and optimize only the pose. This prevents the dynamics components from  explaining translation or large pose changes between consecutive frames. Then, we add the dynamics, $\dyna$, and the dynamics smoothness terms into the optimization for the final optimization of pose and dynamics.

\begin{figure}[t]
    \centerline{
        \includegraphics[height=1.8in]{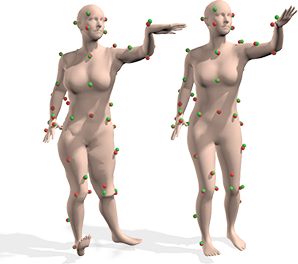} }
    \caption{ Pose estimation with heavy marker occlusion. Pose optimization with constant pose prior weight $\posewt$ (left), variable pose prior weight $\posewt$ (right). $\posewt$ is allowed to vary as a factor of fraction of visible markers resulting in more plausible poses even when toe markers (right foot) and all foot markers (left foot) are missing. Estimated and observed markers are shown in red and green, respectively.}
   
    \label{fig:pose_vary_factor}
    \end{figure}

We explain details of  tuning the weights $\lambda$ in Sec.~\ref{subsec:gridsearch}. 
The velocity constancy weights $\poseConstwt$ and $\dynaConstwt$ depend on the mocap system calibration and optical tracking quality, data frame rate, and the types of motions. Therefore, these values could not be optimized using just one source of data, so we empirically determined them through experiments on different datasets of varying frame rates and motions.
The final weights determined for this stage are:
\begin{multline}
    \datawt=400 \times b , 
    {\posewt}_B=1.6 \times \posecpl, 
    {\posewt}_H=1.0 \times \posecpl,  \\
    \poseConstwt=2.5,  
    \dynawt=1.0, 
    \dynaConstwt=6.0,
\end{multline}
Similar to $b$, which adjusts the weight of the data term to varying markersets, $q$ is a weight-balancing factor for the pose prior $\posewt$. 
During a mocap session, markers may get occluded by the body due to pose. If multiple markers of a particular body part are occluded simultaneously, the optimization may result in unreliable and implausible poses, such as the estimated pose shown in Fig.~\ref{fig:pose_vary_factor} (left). 
    To address this, we introduce a coefficient $\posecpl=1 + \big(\frac{x}{|\Obsmrks|} * 2.5 \big)$, where $x$ is the number of missing markers in a given frame, $|\Obsmrks|$ are the number of total observed markers. This updates the pose prior weight as a factor of the number of missing markers. 
The more markers that are missing, the higher this weights the pose prior.
    This term can increase the prior weight by up to a factor of $\posecpl=3.5$, in the worse case scenario where $x=|\Obsmrks|$, and goes down to having no effect, $\posecpl=1.0$ when all session markers are visible $x=0$. An example of the pose-estimation with this factor is shown in Fig.~\ref{fig:pose_vary_factor} (right).

%% file: sections/07_optimization.tex
\subsection{Optimization and Runtime} 
Similar to MoSh we use Powell’s gradient based dogleg minimization \cite{Nocedal2006NO} implemented in the Chumpy \cite{Loper:2014} auto-differentiation package. Details on the runtime are presented in the Supplementary Material.

%% file: sections/08_evaluation.tex
\section{Evaluation}
\label{sec:evaluation}
In order to set the hyperparameters and evaluate the time varying surface reconstruction results of MoSh++, we need reference ground truth 3D data with variations in shape, pose and soft-tissue deformation.
To that end, we introduce the SSM dataset (Sec. \ref{subsubsec:SSMdata}) %
and optimize the weights of the objective functions (Eqs. \ref{eq:mosh_shape_obj} and \ref{eq:mosh_pose_obj}) using cross-validation on SSM (Sec. \ref{subsec:gridsearch}).
After optimizing the hyper-parameters, we evaluate the accuracy of MoSh++, e.g. 
shape reconstruction accuracy (Sec. \ref{subsec:shape_evaluation}), pose, and soft-tissue motion reconstruction  (Sec. \ref{subsec:pose_evaluation}) on the test set.

\begin{figure}[t]
\centerline{\includegraphics[width=0.8\columnwidth]{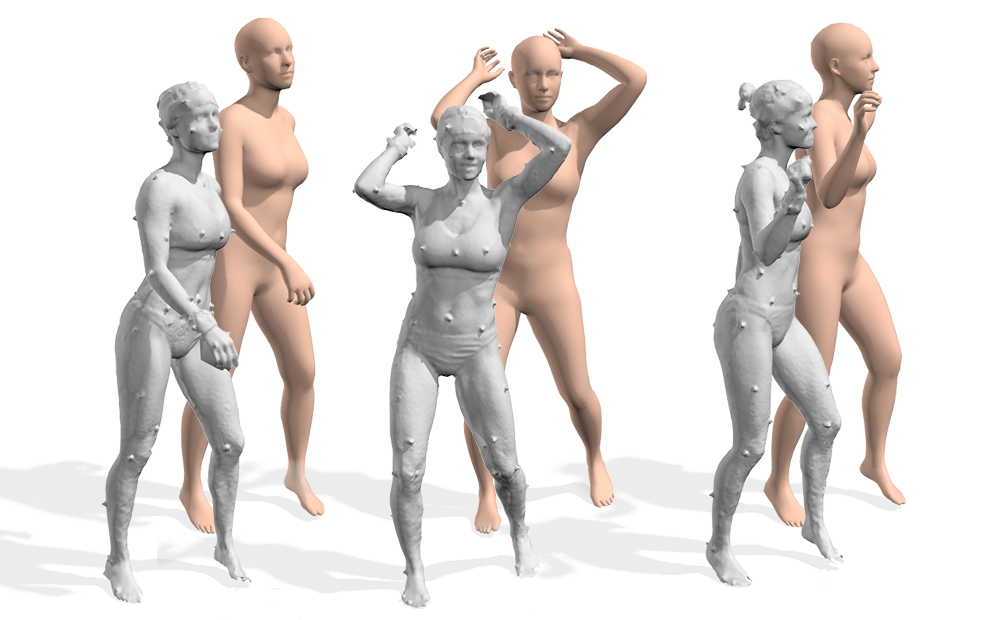}}
\caption{SSM dataset. 3D scans with mocap markers (gray) and fitted bodies (pink).  The average scan to model distance between them is $7.4mm$.}
\label{fig:ssm}
\end{figure}

\subsection{Synchronized Scans and Markers (SSM)}
\label{subsubsec:SSMdata}
We use an OptiTrack mocap system \cite{optitrack} to capture  subjects with 67 markers; i.e.~using the {\em optimized marker-set} proposed by MoSh. 
The system was synchronized to record the mocap data synchronously with a 4D scanning system \cite{3dmdllcatlantaga4d}. 
See Fig.~\ref{fig:ssm}; details are provided in the supplementary material. 
The dataset consists of three subjects with varying body shapes, performing a total of $30$ different motions. 
Two of the three subjects were professional models who signed modeling contracts; this allows us to release their 4D scan data, along with the synchronized mocap data for the research community.

We evaluate the accuracy of MoSh++ using the 67 markers, as well as a more standard 46 marker subset of the 67 markers.
For both  testing and evaluation, we use scan-to-model distances between the 3D scans (our ground truth mesh) of the SSM dataset and the corresponding estimated meshes for each trial of the hyper-parameter search and evaluation.
For each reconstructed mocap frame, we take a uniform sampling of 10,000 points of the corresponding synchronized 3D scan and compute the distance from each of these to the closest surface point in our reconstructed mesh. We measure the average of these distances (in $mm$). 

\subsection{Hyper-parameter Search using \dataset}
    \label{subsec:gridsearch}
The goal is to set the $\lambda$ weights in Eq.~\ref{eq:mosh_shape_obj} and  Eq.~\ref{eq:mosh_pose_obj} to minimize the reconstruction error for the validation data. Grid search complexity grows exponentially with the number of parameters (i.e. $5$ parameters in the case of shape estimation, $4$ in the case of pose estimation). Therefore, we perform line search on each parameter keeping the others fixed. 

For the shape estimation stage, the optimization uses $12$ randomly chosen mocap frames from each training subject to estimate shape and marker location for that subject. 
Instead of choosing a single, unseen pose to evaluate shape accuracy as in~\cite{loper2014mosh}, we report the average error over the $12$ randomly selected frames from the first stage of Mosh (see Sec. \ref{sec:model_fitting}). 
Here the duration of the mocap sessions does not matter but variation of body shape among the testing and training subjects is important. Therefore, we separate mocap data of two out of the three SSM subjects as the training set, while keeping the third subject held out for testing and evaluation. 
We repeat the process $4$ times for the training subjects, %
using a different random set of $12$ frames for each trial. Validation is performed by running the optimization a fifth time, and initializing with a new randomization seed. We use a line search strategy to determine objective weights [$\datawt$, $\posewt$, $\shapewt$, $\initwt$, $\StoMwt$] of Eq. \ref{eq:mosh_shape_obj} by finding a combination of these weights that provide the lowest reconstruction error for the estimated body mesh in the $12$ frames picked during each trial. 
The final weights are described in Sec.~\ref{sec:model_fitting}.

For pose estimation, we separated $20\%$ of the total captured mocap files from the three subjects as a held-out set for testing and evaluation. The first $200$ frames of the rest of the motion files are used for training, leaving the remaining frames (roughly $60\%$ of the training set) for validation. We perform a line search on the objective weights [$\datawt$, $\posewt$, $\dynawt$] of Eq. \ref{eq:mosh_pose_obj} and the missing-marker coefficient $\posecpl$, obtaining the final weights described in Sec.~\ref{sec:model_fitting}. 
\subsection{Shape Estimation Evaluation}
\label{subsec:shape_evaluation}
\begin{figure*}
    \centering{
        \includegraphics[width=\linewidth]{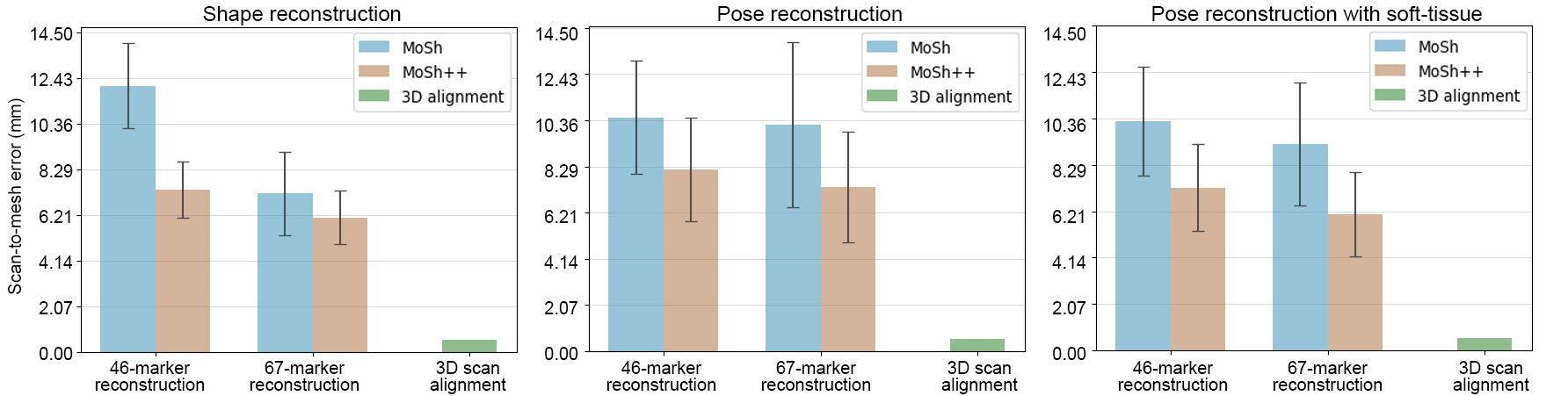}
        }
    \caption{MoSh vs MoSh++ shape and pose reconstruction:
    Mean absolute distance of body shapes reconstructed, using MoSh using the BlendSCAPE model (blue bars) and MoSh++ using SMPL and optimized hyper-parameters (orange bars), to ground-truth 3D scans (green bars), for evaluating 1) Shape estimation, 2) Pose estimation, 3) Pose estimation with DMPL. Error bars indicate standard deviations. We compare a standard 46 marker set with the 67 marker set of MoSh~\cite{loper2014mosh}. MoSh++ with only 46 markers is nearly as good as MoSh with 67 markers.
    Average scan-to-mesh distance between 3D scan alignments and the original scans are shown in green as a baseline for comparison, e.g. an average value of $0.5mm$.}
    \label{fig:moshpp_reconstruction}
\end{figure*}
Compared to MoSh, we obtain more accurate results on $\dataset$.
Fig. \ref{fig:moshpp_reconstruction} (left) shows that the shape estimation accuracy on SSM is $12.1mm$ and $7.4mm$ for MoSh and MoSh++ respectively, when using a standard 46-markerset. 
Note that we use SSM to determine the optimal number of shape and dynamic coefficients (16 and 8 respectively).
Adding more decreases marker error but this overfits to the markers, causing higher error compared with the ground truth shape.
Details are in the Supplemental Material.
\subsection{Pose and Soft-tissue Estimation Evaluation}
\label{subsec:pose_evaluation}
\begin{figure}
\centering{
    \includegraphics[width=3.4in]{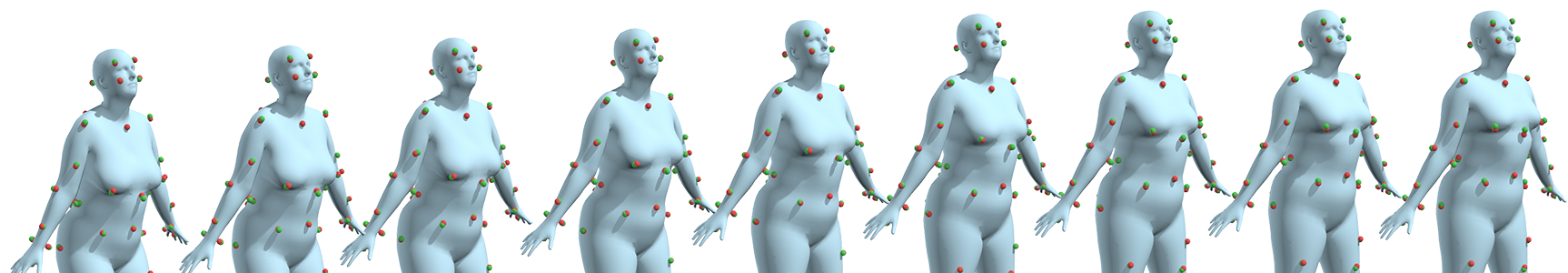}
    \includegraphics[width=3.4in]{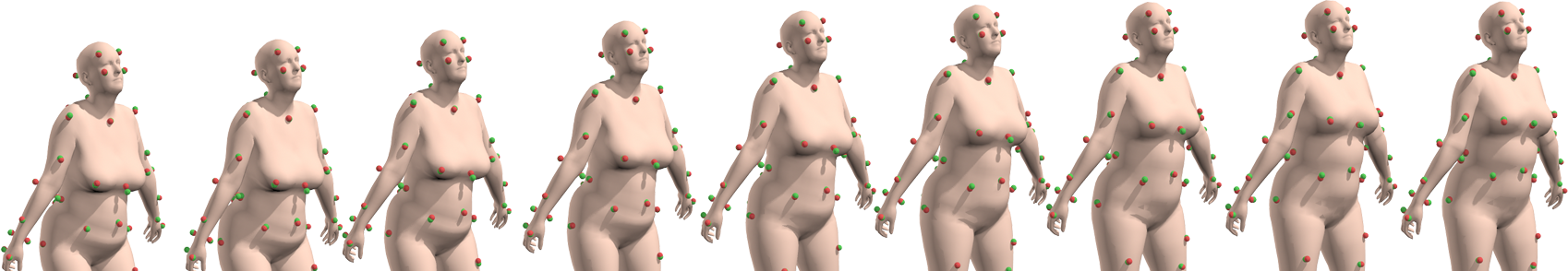}
    \includegraphics[width=3.4in]{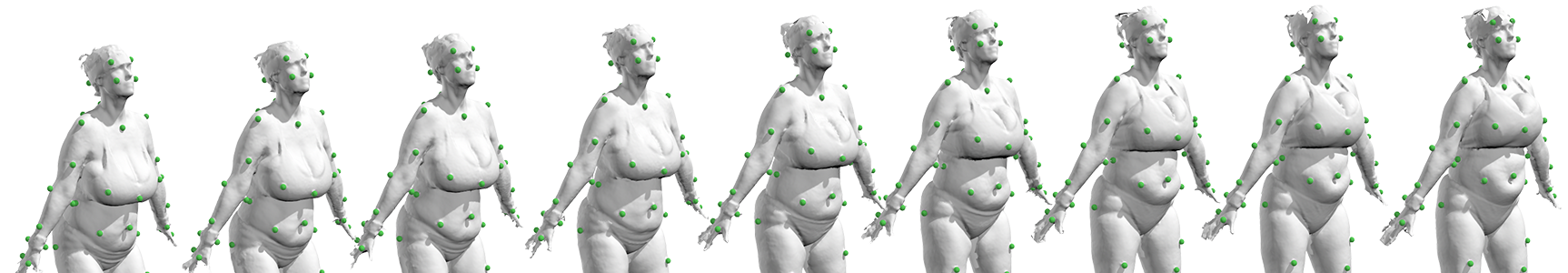} 
    }
\caption{ 
    Soft-tissue Dynamics
    Example animation generated using MoSh~\cite{loper2014mosh} (top row, blue),
    MoSh++ with dynamics from DMPL (second row, orange), 
    and ground truth scans synchronized with Mocap (third row, grey). The estimated marker locations from both, MoSh and MoSh++ are shown on the estimated bodies in red, while the observed mocap markers are shown in green.  MoSh++ captures the soft-tissue motion of the chest and stomach regions much more accurately.
    }
\label{fig:dmpl}
\end{figure}
We also evaluate the per frame accuracy of pose and soft-tissue motion estimation of MoSh++. 
Fig. \ref{fig:moshpp_reconstruction} (middle) shows that
the pose estimation accuracy on SSM without soft-tissue motion estimation is $10.5mm$ and $8.1mm$ for MoSh and MoSh++ respectively, when using a standard 46-markerset. Similarly, with dynamics terms turned-on, MoSh++ achieves more accurate results than MoSh ($7.3mm$ vs $10.24mm$), Figure \ref{fig:moshpp_reconstruction} (right).
The importance of soft-tissue estimation can be observed in Fig.~\ref{fig:dmpl}. 
This result is expected since MoSh \cite{loper2014mosh} models soft-tissue motion in the form of changes in the identity shape space of the BlendSCAPE model, whereas MoSh++ fits the proper DMPL space of soft-tissue motions learned from data \cite{SMPL:2015} to mocap markers.

\subsection{Hand Articulation}
\label{sec:MANO}

    \begin{figure}
    \centering{
        \includegraphics[width=3.1in]{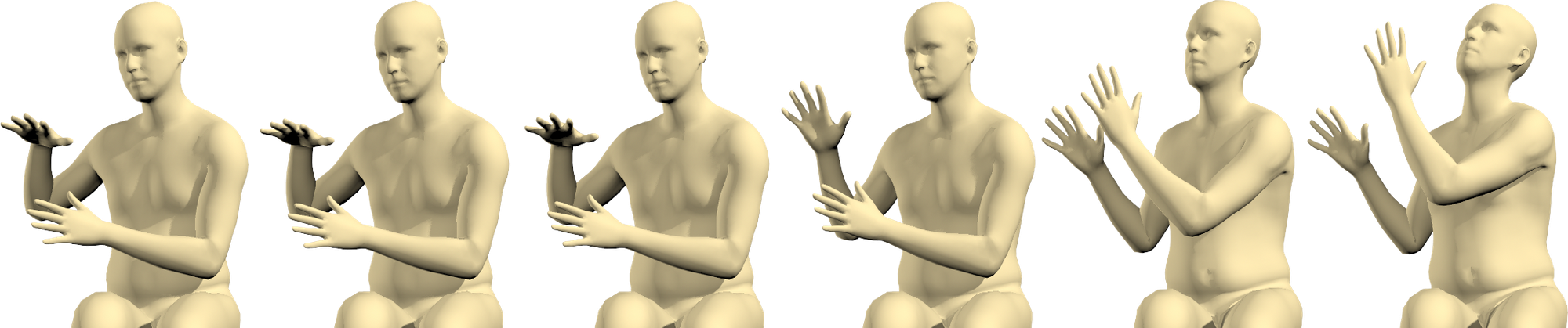}
        \includegraphics[width=3.1in]{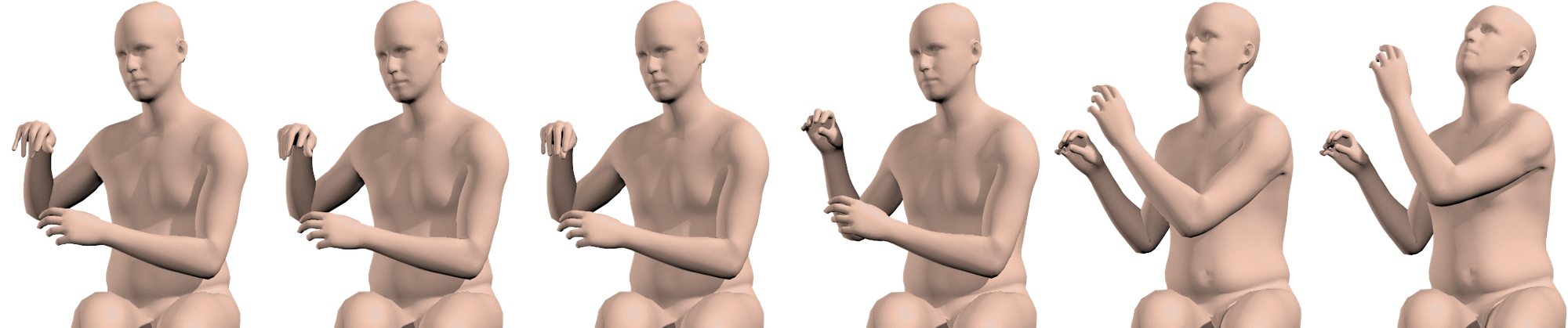} 
        }
    \caption{ 
    Articulated hands:
    \em When hand markers are present, e.g. TCD dataset, MoSh++ fits the hand pose using SMPL-H\cite{MANO:SIGGRAPHASIA:2017}. Top row:  MoSh++ without hands.  Bottom row:  MoSh++ with hand articulation.
    }
    \label{fig:mano}
    \end{figure}
We do not have ground-truth data for evaluating accuracy of hand articulation. %
Qualitative results of our joint body and hand captures can be seen in Fig.~\ref{fig:mano}. Notice how MoSh++ with hand capture leads to more realistic hand poses. 
This illustrates that MoSh++ is not limited to the main body but can be  extended to capture other parts if a model is available.

%% file: sections/09_amass.tex
\section{AMASS Dataset}
\label{sec:AMASS}
We \emph{amassed} in total $\amassNDatasets$ mocap datasets, summarized in Table~\ref{tab:AMASS}.
Each dataset was recorded using a different number of markers placed at different locations on the body; even within a dataset, the number of markers varies.
The publicly available datasets were downloaded from the internet.  
We obtained several other datasets privately or recorded them ourselves (Dancers, Transitions, BioMotionLab and SSM).
We used MoSh++ to map this large amount of marker data into our common
SMPL pose, shape, and soft-tissue parameters.
Due to inherent problems with mocap, such as swapped and mislabeled
markers, we manually inspected the results and either corrected or held out problems.
Fig. \ref{fig:amass_small} shows a few representative examples from different datasets. 
The supplementary material provides video clips that illustrate the diversity and quality of the dataset.
The result is AMASS, the largest public dataset of human shape and pose, including $\amassNSubjExact$ subjects, $\amassNMotionsExact$  motions and $\amassHoursExact$ hours of recordings and is available to the research community at \url{\amassWebAddress}.

\begin{table}[]
		\resizebox{1.\columnwidth}{!}{%
	\begin{tabular}{llrrr}
\textbf{}      & {\footnotesize Markers} &
                                                  {\footnotesize
                                                  Subjects} &
                                                              {\footnotesize
                                                              Motions}
          & {\footnotesize Minutes}  \\
ACCAD~\cite{ACCAD}          & 82               & 20                & 258              & 27.22             \\
BioMotion~\cite{BioMotion}   & 41               & 111               & 3130             & 541.82            \\
CMU~\cite{CMU}          & 41               & 97                & 2030             & 559.18            \\
EKUT~\cite{Mandery2015_KITMotionDB}           & 46               & 4                 & 349              & 30.74             \\
Eyes Japan~\cite{Eyes_Japan}    & 37               & 12                & 795              & 385.42            \\
HumanEva~\cite{HEva}      & 39               & 3                 & 28               & 8.48              \\
KIT~\cite{Mandery2015_KITMotionDB}         & 50               & 55                & 4233             & 662.04            \\
MPI HDM05~\cite{MPI_HDM05}      & 41               & 4                 & 219              & 147.63            \\
MPI Limits~\cite{MPI_limits}     & 53               & 3                 & 40               & 24.14             \\
MPI MoSh~\cite{loper2014mosh}     & 87               & 20                & 78               & 16.65             \\
SFU~\cite{sfu}            & 53               & 7                 & 44               & 15.23             \\
SSM            & 86               & 3                 & 30               & 1.87              \\
TCD Hand~\cite{TCD_hands}      & 91               & 1                 & 62               & 8.05              \\
TotalCapture~\cite{TotalCapture}   & 53               & 5                 & 40               & 43.71             \\
Transitions   & 53               & 1                 & 115              & 15.84             \\
\textbf{Total} & \textbf{-}       & \textbf{346}      & \textbf{11451}   & \textbf{2,488.01}
	\end{tabular}
}
	\caption{Datasets contained in AMASS. 
	We use MoSh++ to map more than $\amassHoursExact$  hours of marker data into SMPL parameters, giving a unified format. %
	}
	\label{tab:AMASS} 
\end{table}

%% file: sections/dl_application.tex
\section{Deep Learning Application}
One application of AMASS is in training deep neural networks to represent human pose.
Our hypothesis is that previous datasets are too small for this task.
To evaluate this, we retrain a recently proposed human body pose prior (VPoser) \cite{SMPLifypp2019} on different subsets of AMASS. 
The pose prior is used for the task of lifting 2D pose to 3D and is evaluated by mean vertex-to-vertex body joint error on the EHF \cite{SMPLifypp2019} dataset. 
For more details on the dataset and the computation of the error we refer the reader to their original paper. 
Here we use the code provided by  \cite{SMPLifypp2019} to train VPoser on three different subsets of AMASS and report evaluation error for each in Table \ref{tab:dl_applications}.
As expected, the more data we use the lower gets the error, suggesting that the pose prior improves with more data.
The trained VPoser models will be released along with AMASS.
\begin{table}[]
	\resizebox{.8\columnwidth}{!}{%
		\begin{tabular}{ll}
			\textit{Dataset Variant} & \textit{v2v error} \\ 	\hline
			CMU                      & 52.27mm            \\
			CMU, H3.6M, MPI Limits   & 51.87mm            \\ 
			AMASS                    & \textbf{XX.XXmm}   \\
		\end{tabular}%
	}
	\caption{AMASS improves a state of the art human pose prior. }
	\label{tab:dl_applications}
\end{table}

%% file: sections/10_future.tex
\section{Future Work and Conclusions}
Future work will extend SSM dataset to include captures with articulated hands. We also intend to extend MoSh++ to work with facial mocap markers. This should be possible using the recently published FLAME model \cite{FLAME:SiggraphAsia2017}, which uses a similar parametrization as the SMPL and MANO models to model facial identity and expressions.
Current runtime for MoSh++ is not real-time. However, in principle it should be possible to improve the runtime of MoSh++ significantly by using a parallel implementation of SMPL using frameworks such as TensorFlow \cite{abadiagarwaletal2016}. 
Finally, we see an opportunity to push our approach further to address the problems of missing markers and to exploit the body for fully automatic marker labeling.
AMASS itself can be leveraged for this task and used to train models that denoise mocap data (cf. \cite{holden2018robust}).

In conclusion, we have introduced MoSh++, which extends MoSh and enables us to unify marker-based motion capture recordings, while being more accurate than simple skeletons or the previous BlendSCAPE version.
This allowed us to collect the AMASS dataset containing more than $\amassHoursApprox$ hours of mocap data in a unified format consisting of SMPL pose (with articulated hands), shape and soft-tissue motion.
We will continuously incorporate more mocap data contributed by us and the community.\\
\\

%% file: sections/11_acks.tex
\section{Acknowledgements}

We thank Iman Abbasnejad for his immense support and assistance during early development stages of this project, 
Dr. Senya Polikovsky, Andrea Keller and Haiwen Feng for help in project coordination, motion capture and 4D scanning and data processing,
and
Talha Zaman, Dr. Javier Romero and Dr. Dimitris Tzionas for their invaluable advice, technical guidance and helpful discussions on body models and optimization techniques.

N. F. Troje held an NSERC Discovery Grant in support of his research and used the Humboldt Research Award granted by the Alexander-von-Humboldt Foundation to fund several visits with his collaborators. 

Gerard Pons-Moll has been funded throughout this work by MPI and by the Deutsche Forschungsgemeinschaft (DFG German Research Foundation) project 409792180.

We are grateful to all the authors that have made their mocap datasets available and allowed us to include them into our database:

\begin{itemize}
    \item ACCAD dataset was was obtained from \url{https://accad.osu.edu/research/motion-lab/system-data}

    \item The Biomotion Lab dataset (BML) belongs to Dr. Nikolaus F. Troje, who is currently at York University, Toronto, Canada.

    \item The CMU mocap data used in this project was obtained from \url{http://mocap.cs.cmu.edu}. That database was created with funding from NSF EIA-0196217. 

    \item CMU Kitchen data was obtained from \url{http://kitchen.cs.cmu.edu} and the data collection was funded in part by the National Science Foundation under Grant No. EEEC-0540865.

    \item Eyes Japan motion capture data is licensed by mocapdata.com, Eyes, JAPAN Co. Ltd. under the Creative Commons Attribution 2.1 Japan License. To view a copy of this license visit \url{http://creativecommons.org/licenses/by/2.1/jp/}.

    \item Human Eva dataset was obtained from \url{http://humaneva.is.tue.mpg.de/}. This project was supported in part by gifts from Honda Research Institute and Intel Corporation.

    \item KIT and EKUT data was obtained from KIT Whole Body Motion Database \url{https://motion-database.humanoids.kit.edu/}

    \item MPI HDM05 data was was obtained from HDM05 \url{http://resources.mpi-inf.mpg.de/HDM05/}

    \item SFU data was was obtained from \url{http://mocap.cs.sfu.ca}. The database was created with funding from NUS AcRF R-252-000-429-133 and SFU President’s Research Start-up Grant.

    \item TCD Hands dataset was provided by Trinity College Dublin. The work was sponsored by Science Foundation Ireland as part of the Captavatar, NaturalMovers and Metropolis projects.

    \item Total Capture dataset was obtained from \url{http://cvssp.org/data/totalcapture/}. The work was supported by an EPSRC doctoral bursary and InnovateUK via the Total Captureproject, grant agreement 102685. The work was supported in part by the Visual Media project(EU H2020 grant 687800) and through donation of GPU hardware by Nvidia corporation.

    \item The Transitions dataset was provided by Dr. Mazen Al Borno, Dr. Javier Romero and Andrea Keller who contributed to the design and capture of this dataset. 

\end{itemize}

%% file: sections/supplementary/optimization.tex
\section{Optimization and Runtime} 
Section 3.3 of the main paper briefly describes the implementation of MoSh++. %
All methods were tested on CPU using an early 2015 edition MacBook Pro with Intel Core i7, running at 3.1GHz and 16GB RAM. 
Runtimes for each step depend heavily on the frame rate and pose variation of the motion. 
For sequences in the SSM dataset, average runtimes of MoSh++ are as follows: 
\begin{itemize}
   	\item Stage I: about $25$ min/motion sequence
   	\item Stage II without dynamics: about $0.5$ sec/frame
   	\item Stage II with dynamics: about $2$ sec/frame
\end{itemize}

%% file: sections/supplementary/ssm_hardware.tex
\section{Data Collection} 
Section.4.1 of the main paper presents the SSM (Synchronized Scans and Markers) dataset. 
To record this dataset we use an optical motion capture system synchronized and calibrated together with a high resolution 4D scanning system. 

We used an OptiTrack motion capture system (NaturalPoint, Inc. DBA OptiTrack. Corvallis, OR) \cite{optitrack} consisting of 24 Optitrack Prime 17W optical mocap cameras. 
Each subject was fitted with 67 reflective mocap markers based on the {\em optimized marker-set} layout proposed in \cite{loper2014mosh}. 
The subjects wore minimal clothing to avoid artefacts due to sliding of cloth. 
The markers were placed directly on the skin of the subjects wherever possible.

The motion capture system was synchronized to be triggered with a 3dMD 4D body scanning system (3dMD LLC, Atlanta, GA) \cite{3dmdllcatlantaga4d}. The 4D scanner is capable of capturing high-resolution 3D scans of a person at 60 frames per second. The 4D system uses 22 pairs of stereo cameras, 22 color cameras and 34 speckle projectors and arrays of white-light LED panels.

%% file: sections/supplementary/model_size.tex
\begin{figure}[t]
	\centering{
		\includegraphics[width=\columnwidth]{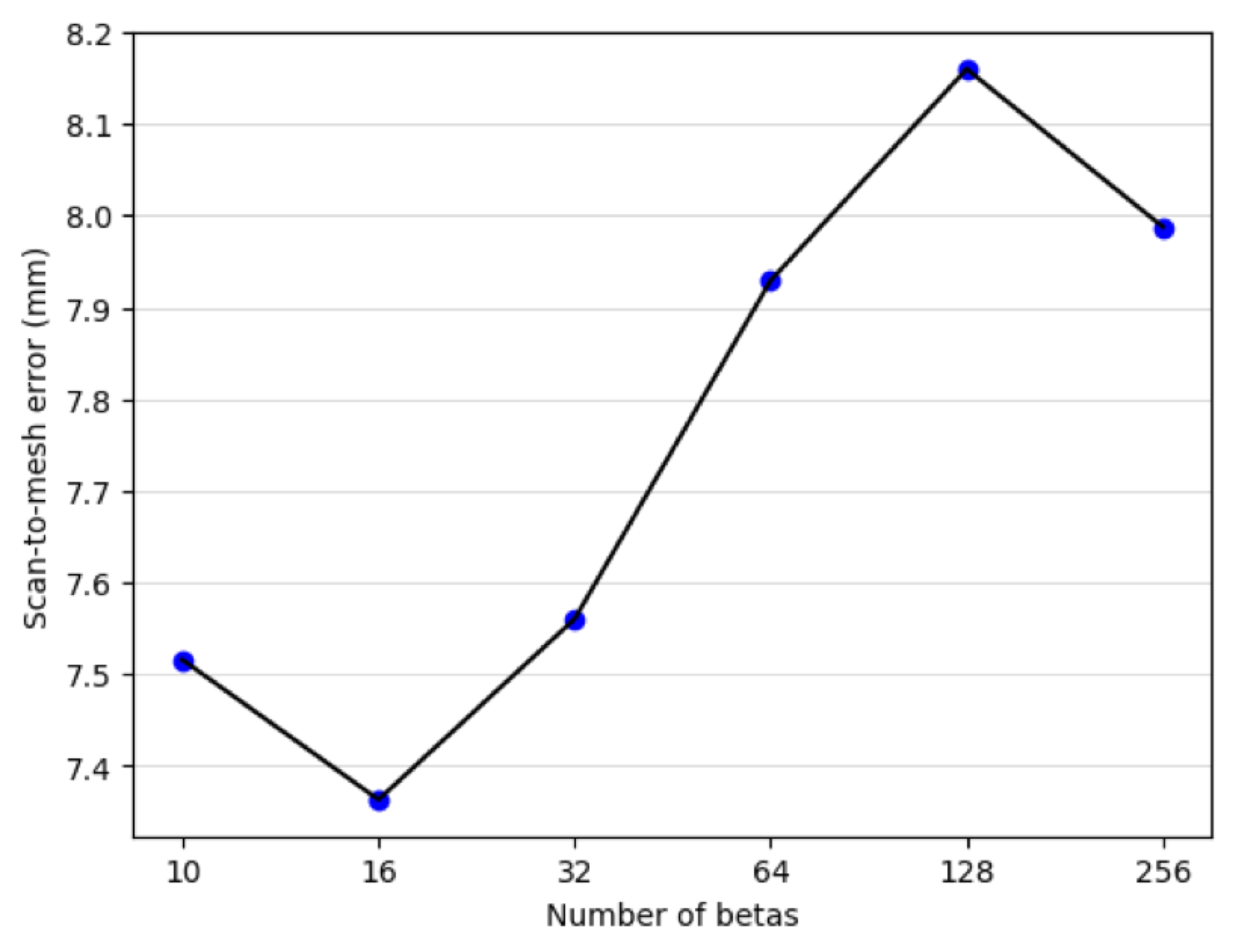}\\
		\includegraphics[width=\columnwidth]{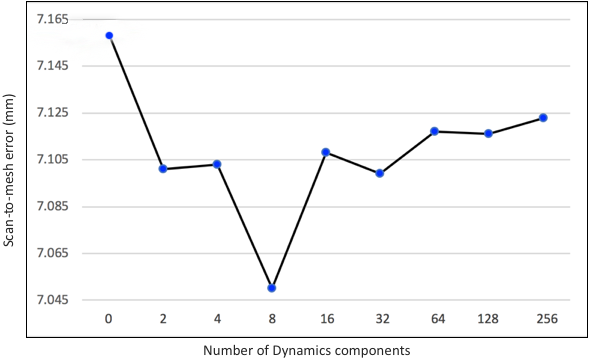}
	}
	\caption{Optimal number of shape and dynamics components. Mesh reconstruction errors on the SSM dataset using varying numbers of SMPL shape components $\shape$ (top), and DMPL dynamic components $\dyna$ (bottom) to find the optimal number to use for shape and soft-tissue optimization. }
	\label{fig:num_smpls_dmpls}
\end{figure}

\section{Model Size}
\label{sup:comp_search}
Section 3.1 of the main paper describes the SMPL body model
incorporated in the MoSh++ pipeline. 
We experimented with varying numbers of SMPL shape components and DMPL
dynamic components to find the optimal number to use to capture shape and
soft-tissue motion. 
We found that $|\shape| = 16$, and $|\dyna|=8$ do the best job of
minimizing error on the held-out validation set and also produce
natural looking soft-tissue deformations. 
Given a limited number of mocap markers, allowing more shape variation
results in overfitting (see Fig.~\ref{fig:num_smpls_dmpls}).

%% file: sections/supplementary/video_samples.tex
\section{Diversity and Quality}
We provide a video to illustrate the variations in the motions in
AMASS and the
quality of reconstructed body surface deformations from the mocap
markers.
Please see the video.  
Note that AMASS captures a wide range of body shapes and motions.

%% file: amass.bbl
\begin{thebibliography}{10}\itemsep=-1pt

\bibitem{3dmdllcatlantaga4d}
{{4D Scan}}.
\newblock {\em http://www.3dmd.com/}.

\bibitem{sfu}
{{SFU Motion Capture Database}}.
\newblock {\em http://mocap.cs.sfu.ca/}.

\bibitem{MocapClub}
{{MocapClub}}.
\newblock {\em http://www.mocapclub.com/}, 2007.

\bibitem{CMUKitchen}
{{CMUKitchen}}.
\newblock {\em http://kitchen.cs.cmu.edu/}, 2009.

\bibitem{ACCAD}
{{ACCAD}}.
\newblock {\em https://accad.osu.edu/research/motion-lab/system-data}, 2018.

\bibitem{Eyes_Japan}
Eyes {{Japan}}.
\newblock {\em http://mocapdata.com}, 2018.

\bibitem{abadiagarwaletal2016}
M.~Abadi, A.~Agarwal, P.~Barham, E.~Brevdo, Z.~Chen, C.~Citro, G.~S. Corrado,
  A.~Davis, J.~Dean, M.~Devin, S.~Ghemawat, I.~Goodfellow, A.~Harp, G.~Irving,
  M.~Isard, Y.~Jia, R.~Jozefowicz, L.~Kaiser, M.~Kudlur, J.~Levenberg, D.~Mane,
  R.~Monga, S.~Moore, D.~Murray, C.~Olah, M.~Schuster, J.~Shlens, B.~Steiner,
  I.~Sutskever, K.~Talwar, P.~Tucker, V.~Vanhoucke, V.~Vasudevan, F.~Viegas,
  O.~Vinyals, P.~Warden, M.~Wattenberg, M.~Wicke, Y.~Yu, and X.~Zheng.
\newblock {{TensorFlow}}: {{Large}}-{{Scale Machine Learning}} on
  {{Heterogeneous Distributed Systems}}.
\newblock {\em arXiv:1603.04467 [cs]}, 2016.

\bibitem{MPI_limits}
I.~Akhter and M.~J. Black.
\newblock Pose-{{Conditioned Joint Angle Limits}} for {{3D Human Pose
  Reconstruction}}.
\newblock In {\em {{IEEE Conf}}.~on {{Computer Vision}} and {{Pattern
  Recognition}} ({{CVPR}} 2015)}, 2015.

\bibitem{alexanderson2016robust}
S.~Alexanderson, C.~O'Sullivan, and J.~Beskow.
\newblock Robust online motion capture labeling of finger markers.
\newblock In {\em Proceedings of the 9th {{International Conference}} on
  {{Motion}} in {{Games}}}. {ACM}, 2016.

\bibitem{allen2003space}
B.~Allen, B.~Curless, and Z.~Popovi\'c.
\newblock The space of human body shapes: {{Reconstruction}} and
  parameterization from range scans.
\newblock {\em ACM Transactions on Graphics (TOG)}, 2003.

\bibitem{Allen:2006:LCM}
B.~Allen, B.~Curless, Z.~Popovi\'c, and A.~Hertzmann.
\newblock Learning a {{Correlated Model}} of {{Identity}} and
  {{Pose}}-dependent {{Body Shape Variation}} for {{Real}}-time {{Synthesis}}.
\newblock In {\em Proceedings of the 2006 {{ACM SIGGRAPH}}/{{Eurographics
  Symposium}} on {{Computer Animation}}}, {{SCA}} '06, 2006.

\bibitem{ANDRIACCHI20001217}
T.~P. Andriacchi and E.~J. Alexander.
\newblock Studies of human locomotion: Past, present and future.
\newblock {\em Journal of Biomechanics}, 2000.

\bibitem{Anguelov05}
D.~Anguelov, P.~Srinivasan, D.~Koller, S.~Thrun, J.~Rodgers, and J.~Davis.
\newblock {{SCAPE}}: {{Shape Completion}} and {{Animation}} of {{PEople}}.
\newblock {\em ACM Transactions on Graphics}, 2005.

\bibitem{CMU}
F.~{De la Torre}, J.~Hodgins, A.~Bargteil, X.~Martin, J.~Macey, A.~Collado, and
  P.~Beltran.
\newblock Guide to the carnegie mellon university multimodal activity
  (cmu-mmac) database.
\newblock {\em Robotics Institute}, 2008.

\bibitem{dueck1990threshold}
G.~Dueck and T.~Scheuer.
\newblock Threshold accepting: A general purpose optimization algorithm
  appearing superior to simulated annealing.
\newblock {\em Journal of computational physics}, 1990.

\bibitem{forney1973viterbi}
G.~D. Forney.
\newblock The viterbi algorithm.
\newblock {\em Proceedings of the IEEE}, 1973.

\bibitem{gorton2009assessment}
G.~E. Gorton, D.~A. Hebert, and M.~E. Gannotti.
\newblock Assessment of the kinematic variability among 12 motion analysis
  laboratories.
\newblock {\em Gait \& posture}, 2009.

\bibitem{Han:2018:OOM}
S.~Han, B.~Liu, R.~Wang, Y.~Ye, C.~D. Twigg, and K.~Kin.
\newblock Online {{Optical Marker}}-based {{Hand Tracking}} with {{Deep
  Labels}}.
\newblock {\em ACM Trans. Graph.}, 2018.

\bibitem{hirshberg2012coregistration}
D.~A. Hirshberg, M.~Loper, E.~Rachlin, and M.~J. Black.
\newblock Coregistration: {{Simultaneous}} alignment and modeling of
  articulated {{3D}} shape.
\newblock In A.~Fitzgibbon, S.~Lazebnik, P.~Perona, Y.~Sato, and C.~Schmid,
  editors, {\em European {{Conference}} on {{Computer Vision}}}, Lecture
  {{Notes}} in {{Computer Science}}. {Springer}, {Springer Berlin Heidelberg},
  2012.

\bibitem{holden2018robust}
D.~Holden.
\newblock Robust {{Solving}} of {{Optical Motion Capture Data}} by
  {{Denoising}}.
\newblock {\em ACM Transactions on Graphics}, 2018.

\bibitem{holden2016deep}
D.~Holden, J.~Saito, and T.~Komura.
\newblock A deep learning framework for character motion synthesis and editing.
\newblock {\em ACM Transactions on Graphics (TOG)}, 2016.

\bibitem{TCD_hands}
L.~Hoyet, K.~Ryall, R.~McDonnell, and C.~O'Sullivan.
\newblock Sleight of {{Hand}}: {{Perception}} of {{Finger Motion}} from
  {{Reduced Marker Sets}}.
\newblock In {\em Proceedings of the {{ACM SIGGRAPH Symposium}} on
  {{Interactive 3D Graphics}} and {{Games}}}, {{I3D}} '12. {ACM}, 2012.

\bibitem{lafortune1992three}
M.~Lafortune, P.~Cavanagh, H.~Sommer, and A.~Kalenak.
\newblock Three-dimensional kinematics of the human knee during walking.
\newblock {\em Journal of biomechanics}, 1992.

\bibitem{leardini2005human}
A.~Leardini, L.~Chiari, U.~D. Croce, and A.~Cappozzo.
\newblock Human movement analysis using stereophotogrammetry: {{Part}} 3.
  {{Soft}} tissue artifact assessment and compensation.
\newblock {\em Gait \&/ Posture}, 2005.

\bibitem{FLAME:SiggraphAsia2017}
T.~Li, T.~Bolkart, M.~J. Black, H.~Li, and J.~Romero.
\newblock Learning a model of facial shape and expression from {{4D}} scans.
\newblock {\em ACM Transactions on Graphics, (Proc. SIGGRAPH Asia)}, 2017.

\bibitem{Loper:2014}
M.~Loper.
\newblock Chumpy.
\newblock 2013.

\bibitem{loper2014mosh}
M.~Loper, N.~Mahmood, and M.~J. Black.
\newblock {{MoSh}}: {{Motion}} and shape capture from sparse markers.
\newblock {\em ACM Transactions on Graphics (TOG)}, 2014.

\bibitem{SMPL:2015}
M.~Loper, N.~Mahmood, J.~Romero, G.~{Pons-Moll}, and M.~J. Black.
\newblock {{SMPL}}: {{A Skinned Multi}}-{{Person Linear Model}}.
\newblock {\em ACM Trans. Graphics (Proc. SIGGRAPH Asia)}, 2015.

\bibitem{Mandery2015_KITMotionDB}
C.~Mandery, O.~Terlemez, M.~Do, N.~Vahrenkamp, and T.~Asfour.
\newblock The {{KIT Whole}}-{{Body Human Motion Database}}.
\newblock In {\em International {{Conference}} on {{Advanced Robotics}}
  ({{ICAR}})}, 2015.

\bibitem{maycock2015fully}
J.~Maycock, T.~Rohlig, M.~Schroder, M.~Botsch, and H.~Ritter.
\newblock Fully automatic optical motion tracking using an inverse kinematics
  approach.
\newblock In {\em Humanoid {{Robots}} ({{Humanoids}}), 2015 {{IEEE}}-{{RAS}}
  15th {{International Conference}} On}. {IEEE}, 2015.

\bibitem{Mueller_09}
M.~M\"uller, A.~Baak, and H.-P. Seidel.
\newblock Efficient and {{Robust Annotation}} of {{Motion Capture Data}}.
\newblock In {\em Proceedings of the {{ACM SIGGRAPH}}/{{Eurographics
  Symposium}} on {{Computer Animation}} ({{SCA}})}, 2009.

\bibitem{MPI_HDM05}
M.~M\"uller, T.~R\"oder, M.~Clausen, B.~Eberhardt, B.~Kr\"uger, and A.~Weber.
\newblock Documentation {{Mocap Database HDM05}}.
\newblock Technical report, {Universit\"at Bonn}, 2007.

\bibitem{optitrack}
I.~NaturalPoint.
\newblock Motion {{Capture Systems}}.

\bibitem{Nocedal2006NO}
J.~Nocedal and S.~J. Wright.
\newblock {\em Numerical {{Optimization}}}.
\newblock {Springer}, New York, 2nd edition, 2006.

\bibitem{Dyna:SIGGRAPH:2015}
G.~{Pons-Moll}, J.~Romero, N.~Mahmood, and M.~J. Black.
\newblock Dyna: {{A Model}} of {{Dynamic Human Shape}} in {{Motion}}.
\newblock {\em ACM Transactions on Graphics, (Proc. SIGGRAPH)}, 2015.

\bibitem{MANO:SIGGRAPHASIA:2017}
J.~Romero, D.~Tzionas, and M.~J. Black.
\newblock Embodied {{Hands}}: {{Modeling}} and {{Capturing Hands}} and {{Bodies
  Together}}.
\newblock {\em ACM Transactions on Graphics, (Proc. SIGGRAPH Asia)}, 2017.
\newblock () Two first authors contributed equally.

\bibitem{schroder2015reduced}
M.~Schr\"oder, J.~Maycock, and M.~Botsch.
\newblock Reduced marker layouts for optical motion capture of hands.
\newblock In {\em Proceedings of the 8th {{ACM SIGGRAPH Conference}} on
  {{Motion}} in {{Games}}}. {ACM}, 2015.

\bibitem{HEva}
L.~Sigal, A.~Balan, and M.~J. Black.
\newblock {{HumanEva}}: {{Synchronized}} video and motion capture dataset and
  baseline algorithm for evaluation of articulated human motion.
\newblock {\em International Journal of Computer Vision}, 2010.

\bibitem{taylor2016efficient}
J.~Taylor, L.~Bordeaux, T.~Cashman, B.~Corish, C.~Keskin, T.~Sharp, E.~Soto,
  D.~Sweeney, J.~Valentin, B.~Luff, et~al.
\newblock Efficient and precise interactive hand tracking through joint,
  continuous optimization of pose and correspondences.
\newblock {\em ACM Transactions on Graphics (TOG)}, 2016.

\bibitem{tkach2017online}
A.~Tkach, A.~Tagliasacchi, E.~Remelli, M.~Pauly, and A.~Fitzgibbon.
\newblock Online generative model personalization for hand tracking.
\newblock {\em ACM Transactions on Graphics (TOG)}, 2017.

\bibitem{BioMotion}
N.~F. Troje.
\newblock Decomposing biological motion: {{A}} framework for analysis and
  synthesis of human gait patterns.
\newblock {\em Journal of Vision}, 2002.

\bibitem{TotalCapture}
M.~Trumble, A.~Gilbert, C.~Malleson, A.~Hilton, and J.~Collomosse.
\newblock Total {{Capture}}: {{3D Human Pose Estimation Fusing Video}} and
  {{Inertial Sensors}}.
\newblock In {\em {{BMVC17}}}, 2017.

\bibitem{varol17}
G.~Varol, J.~Romero, X.~Martin, N.~Mahmood, M.~J. Black, I.~Laptev, and
  C.~Schmid.
\newblock Learning from {{Synthetic Humans}}.
\newblock In {\em {{CVPR}}}, 2017.

\bibitem{Gymnasts}
C.~{von La\ss{}berg}, W.~Rapp, B.~Mohler, and J.~Krug.
\newblock Neuromuscular onset succession of high level gymnasts during dynamic
  leg acceleration phases on high bar.
\newblock {\em Journal of Electromyography and Kinesiology}, 2013.

\end{thebibliography}
